\documentclass[10pt,twocolumn,letterpaper]{article}

\usepackage{iccv}  
\usepackage{enumitem}

\definecolor{iccvblue}{rgb}{0.21,0.49,0.74}
\usepackage[pagebackref,breaklinks,colorlinks,allcolors=iccvblue]{hyperref}

\def\paperID{3643} 
\def\confName{ICCV}
\def\confYear{2025}

\title{Socratic Chart: Cooperating Multiple Agents for Robust SVG Chart Understanding}

\author{Yuyang Ji\\
University of Illinois at Urbana-Champaign\\
{\tt\small yuyangji@illinois.edu}
\and
Haohan Wang\\
University of Illinois at Urbana-Champaign\\
{\tt\small haohanw@illinois.edu}
}

\begin{document}
\maketitle
\begin{abstract}
Multimodal Large Language Models (MLLMs) have shown remarkable versatility but face challenges in demonstrating true visual understanding, particularly in chart reasoning tasks. 
Existing benchmarks like ChartQA reveal significant reliance on text-based shortcuts and probabilistic pattern-matching rather than genuine visual reasoning. 
To rigorously evaluate visual reasoning, we introduce a more challenging test scenario by removing textual labels and introducing chart perturbations in the ChartQA dataset. Under these conditions, models like GPT-4o and Gemini-2.0 Pro experience up to a 30\% performance drop, underscoring their limitations.
To address these challenges, we propose Socratic Chart, a new framework that transforms chart images into Scalable Vector Graphics (SVG) representations, enabling MLLMs to integrate textual and visual modalities for enhanced chart understanding.
Socratic Chart employs a multi-agent pipeline with specialized agent-generators to extract primitive chart attributes (e.g., bar heights, line coordinates) and an agent-critic to validate results, ensuring high-fidelity symbolic representations. 
Our framework surpasses state-of-the-art models in accurately capturing chart primitives and improving reasoning performance, establishing a robust pathway for advancing MLLM visual understanding.
\end{abstract}    
\section{Introduction}
\label{sec:intro}

Multimodal Large Language Models (MLLMs)~\cite{achiam2023gpt, team2023gemini, alayrac2022flamingo, bai2023qwen} have demonstrated remarkable versatility and effectiveness across a wide range of real-world applications. These models integrate vision and language capabilities, enabling them to process and interpret images alongside text, which has led to significant advancements in tasks requiring multimodal understanding. Recent iterations like GPT-4V, Gemini-2.0, and others have shown impressive performance on benchmarks that test visual-linguistic reasoning.

\begin{figure}[t]
\includegraphics[width=0.45\textwidth]{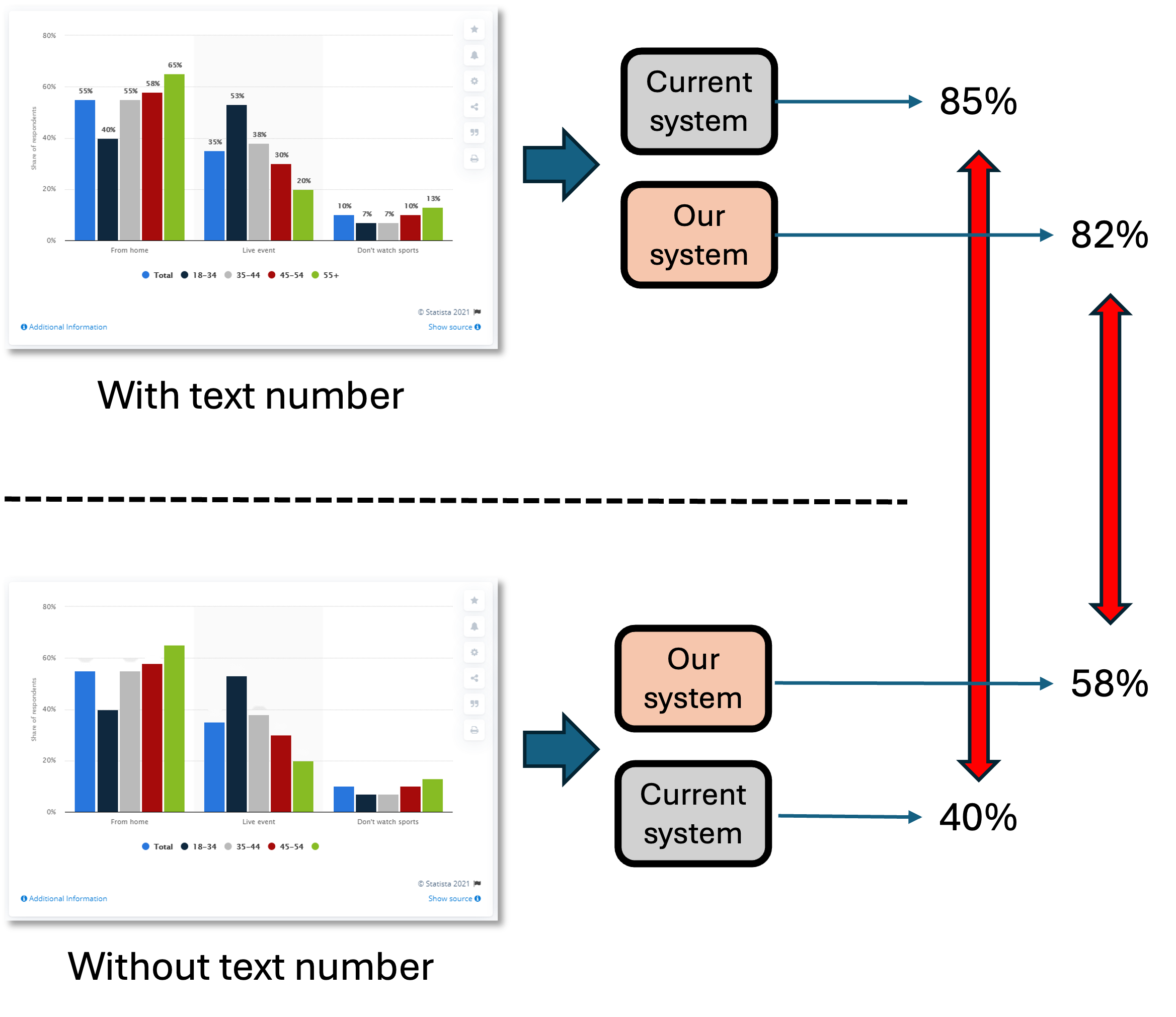}
\caption{Performance comparison of previous system on the ChartQA benchmark under two conditions: with original charts (top) and with charts where text labels have been removed (bottom). Removing textual labels eliminates shortcuts that allow models to rely on OCR-based extraction rather than genuine visual reasoning. The significant performance drops observed in these systems like GPT-4v, Gemini2, and SIMPLOT highlight their dependence on these text-based shortcuts when answering questions. In contrast, our proposed method, \textit{Socratic Chart}, demonstrates a substantially smaller performance drop (23.9\%) due to its innovative framework that transforms chart images into Scalable Vector Graphics (SVG) representations. This approach enables multimodal large language models (MLLMs) to more effectively integrate visual information with textual context, resulting in enhanced chart understanding even when explicit textual cues are absent.
}
\label{fig:comparison}
\vspace{-15pt}
\end{figure}

Among these applications, chart understanding is particularly valuable~\cite{masry2023unichart, liu2022deplot, kim2024simplot}, given the prevalence of charts in scientific papers, articles, and websites. This task poses unique challenges, requiring models to reason over numerical data, textual labels, and intricate visual structures. 
Effective chart comprehension demands precise interpretation and measurement—for example, accurately assessing the height of a bar in a bar chart. Consequently, chart understanding serves as a crucial benchmark for measuring advancements in MLLM capabilities.

However, whether these models possess true visual understanding remains a critical area of research. 
As chartQA has emerged as a popular benchmark for assessing these capabilities~\cite{masry2022chartqa}, the widespread use of it increases the risk of inadvertent data contamination. 
Additionally, the presence of textual labels on geometric elements (e.g., bars) introduces shortcuts, allowing models to rely on OCR to extract key information, such as the height of a bar, rather than leveraging genuine visual reasoning.

Recent efforts such as ChartXiv~\cite{wang2024charxiv} have introduced novel benchmarks to evaluate chart reasoning capabilities more comprehensively, which further highlights the importance of robust chart understanding approaches.
While some studies showcase their impressive capabilities, concerns persist, primarily due to two key observations: (1) many answers can be inferred solely from the textual questions and options, often due to data contamination~\cite{chen2024we,hegde2023analyzing}; and (2) much of the knowledge exhibited by MLLMs may be memorized from their extensive Internet-scale training datasets. Consequently, the reasoning processes in MLLMs often rely on probabilistic pattern-matching rather than genuine formal reasoning~\cite{chen2024we, jiang2024peek}.

To gain deeper insights into the strengths and weaknesses of MLLMs in chart understanding tasks, we introduced two interventions. 
First, we removed the textual labels from the chart dataset to assess the extent to which models rely on text-based information instead of visual capabilities to answer questions. Second, we applied simple perturbations by expanding the charts horizontally or vertically to evaluate whether these modifications affect the models' visual reasoning abilities.
Our findings reveal that these perturbations in the ChartQA dataset can lead to performance deterioration of up to 30\%, highlighting that current MLLMs still struggle with achieving true visual understanding in chart reasoning tasks.

Therefore, methods that can understand the semantics of charts in a more robust manner is needed. 
Thus, we propose a novel method, \textit{Socratic Chart}, which converts chart images into textual representations using Scalable Vector Graphics (SVG). The generated SVG aims to facilitate precise chart comprehension by MLLMs, leveraging textual descriptions alongside visual information. By employing text to describe chart details, Socratic Chart encourages MLLMs to improve their understanding of charts through both modalities.

The SVG text representation provides a high-level abstraction of primitive attributes, enabling MLLMs to interpret and generalize across reasoning tasks in a zero-shot manner. Our method employs a multi-agent collaboration pipeline, dividing the task into specialized subtasks handled by dedicated agent-generators. These agents extract key attributes such as line coordinates, bar heights, and pie chart proportions. Once the agent-generators complete their respective subtasks, an agent-critic performs a deliberative evaluation, refining and validating the outputs to ensure high fidelity in the symbolic representations.

The resulting textual representations, along with the original chart images, are then integrated into MLLMs for inference. Evaluation on ChartQA and related benchmarks show that our framework outperforms state-of-the-art vision-language models (e.g., GPT-4V and Gemini-2.0 Pro). Socratic Chart excels in accurately capturing primitive attributes and improving reasoning performance in chart-based tasks.

We summarize our contribution as follows: 
\begin{itemize}[nosep, leftmargin=0pt, topsep=0pt, partopsep=0pt]
    \item We introduced simple modifications of existing benchmark by (1) removing textual labels and (2) applying horizontal and vertical perturbations for an enhanced ChartQA data for testing the model's true reasoning abilities. 
    Using the enhanced ChartQA dataset, we demonstrated that current MLLMs rely heavily on text-based reasoning, with performance degrading by up to 30\% under interventions, revealing challenges in true visual understanding.
    \item We introduced a novel framework that converts chart images into SVG representations, enabling high-level abstraction of primitive attributes for enhanced MLLM interpretation. Specifically, we developed a multi-agent pipeline with:
    Specialized agent-generators for extracting chart-specific attributes. An agent-critic mechanism to evaluate and refine generated results, ensuring high fidelity in symbolic representations.
\end{itemize}

\section{Related Work}
\label{sec:related_work}

\subsection{Visual Agent and Visual Programming} With the advancement of MLLMs, researchers have explored the potential of breaking down complex vision tasks into simpler substeps, each solvable using dedicated vision tools~\cite{hu2022promptcap, hu2024visual}. Among these approaches, the most relevant to our work are VisProg~\cite{gupta2023visual} and ViperGPT~\cite{suris2023vipergpt}, which leverage MLLMs to generate Python code that sequentially invokes specialized vision tools. Furthermore, recent studies have begun conceptualizing MLLMs as agents capable of both reasoning and taking action~\cite{yao2023react, yasunaga2022retrieval}. This paradigm has been applied across various domains, including robotics~\cite{nasiriany2024pivot}, vision~\cite{yang2023mm}, and GUI navigation~\cite{yan2023gpt}.

\subsection{Multi-modal Reasoning on Scientific Chart} 
Recent works leverage the architecture of multimodal large language models and enhance chart comprehension through supervised fine-tuning on extensive chart instruction datasets. Some approaches adapt vision-language models for chart-related tasks, while others develop plugins that enable LLMs to interpret charts more effectively. Matcha~\cite{liu2022matcha} builds upon Pix2Struct~\cite{lee2023pix2struct} by incorporating mathematical reasoning and chart data extraction, demonstrating strong performance in chart question answering and summarization. Meanwhile, Unichart~\cite{masry2023unichart} undergoes multitask instruction tuning to support a broader range of chart-related tasks, positioning it as the most versatile and effective chart vision-language model available.

\subsection{Scalable Vector Graphics (SVG)}
Scalable Vector Graphics (SVG) is a vector-based image format that encodes visual content using XML to represent elements such as polygons, circles, and rectangles~\cite{peng2004roles}. Unlike raster graphics composed of pixel arrays, vector representations rely on collections of parameterized shapes, each defined by coordinate sets and associated color attributes~\cite{ferraiolo2000scalable}. This approach produces compact, infinitely scalable imagery that maintains sharpness at any zoom level, and is natively supported by modern web browsers without requiring specialized software or plugins. Such flexibility facilitates easy adjustments to properties like stroke width or color, making vector graphics well-suited for web-based and design-oriented applications.

The XML-based structure of SVG files also encodes rich semantic information, including spatial relationships between objects, thereby enabling programmatic manipulation and high-level interpretation. Related efforts have further investigated the bridge between pixel-oriented images and vector formats, exploring the use of advanced vision-language models (VLMs) to ensure that the generated SVGs are semantically coherent and visually succinct~\cite{zhang2023beyond}. These methods demonstrate that leveraging the direct representation of graphical elements in SVG and integrating advanced techniques, such as image-to-SVG conversion supported by large language models, can yield simpler, more interpretable, and semantically coherent vector graphics, which ultimately align better with human readability and reasoning tasks.

\subsection{Benchmarks for Visual Reasoning Capabilities} Several datasets have been introduced to assess the visual perception capabilities of MLLMs on general figures. While some datasets focus on evaluating MLLMs for specific visual perception tasks, such as OCR~\cite{chen2024mmr}, depth estimation~\cite{fu2024blink}, and object counting~\cite{jain2024vcoder}, widely used benchmarks for visual perception in general images~\cite{antol2015vqa, li2024seed} tend to emphasize broader scene understanding rather than fine-grained visual details. This limitation likely arises from the challenge of formulating questions that probe detailed information in general images.
In contrast, scientific figures offer a structured format that facilitates the annotation of objective questions about detailed visual content. Several studies have specifically evaluated MLLMs' visual perception on scientific figures. HallusionBench~\cite{guan2024hallusionbench} reveals that MLLMs struggle with misleading figures, such as those containing illusory geometric shapes. BlindTest~\cite{rahmanzadehgervi2024vision} evaluates the visual perception of four MLLMs, GPT-4o, Gemini-1.5 Pro, Claude-3 Sonnet, and Claude-3.5 Sonnet, using seven simple 2D geometric tasks inspired by human visual acuity tests. These tasks assess basic vision without requiring reasoning or world knowledge.
\section{Method}
\label{sec:method}

In our method, we leverage the power of SVG representations for chart understanding tasks. SVG (Scalable Vector Graphics) serves as the cornerstone of our approach, providing a structured, interpretable representation that bridges the gap between visual chart elements and machine-readable data. To generate these SVG representations from chart images, we design a multi-agent pipeline system that converts chart images into high-fidelity SVG. The generated SVG, along with the original chart image and query, is then provided to MLLMs for chart understanding tasks.

\subsection{SVG Representation for Chart Understanding}

The Chart SVG text representation serves as a high-level abstraction of chart elements, enabling MLLMs to perform reasoning tasks in a zero-shot manner. SVG preserves both geometric and semantic details of charts, maintaining spatial relationships and quantitative data in a structured format accessible to language models:

\begin{verbatim}
<svg width="W" height="H">
  <!-- Bars, lines, pie segments -->
  <!-- Text annotations -->
</svg>
\end{verbatim}

SVG encoding provides several key advantages for chart understanding:

\begin{itemize}[nosep, leftmargin=0pt, topsep=0pt, partopsep=0pt]
    \item \textbf{Structured Representation:} Each chart element is represented as a distinct SVG entity with specific attributes. Bars are encoded as \texttt{<rect>} elements, lines as \texttt{<path>}, pie segments as proportional arcs, and text as \texttt{<text>} elements.
    \item \textbf{Preserved Spatial Relationships:} The SVG format maintains precise positioning of all chart elements, ensuring that spatial relationships between data points, axes, and labels are accurately captured.
    \item \textbf{Semantic Integration:} Text elements are associated with their corresponding visual components, preserving the semantic relationships essential for chart interpretation.
    \item \textbf{Quantitative Fidelity:} Geometric attributes in SVG directly correlate to the underlying data values, allowing for accurate quantitative reasoning.
\end{itemize}

This structured SVG representation enables MLLMs to interpret charts in a zero-shot manner, facilitating complex reasoning tasks such as trend analysis, comparison, and data interpretation without requiring task-specific fine-tuning.

\begin{figure*}[t]
\centering
\includegraphics[width=1.0\textwidth]{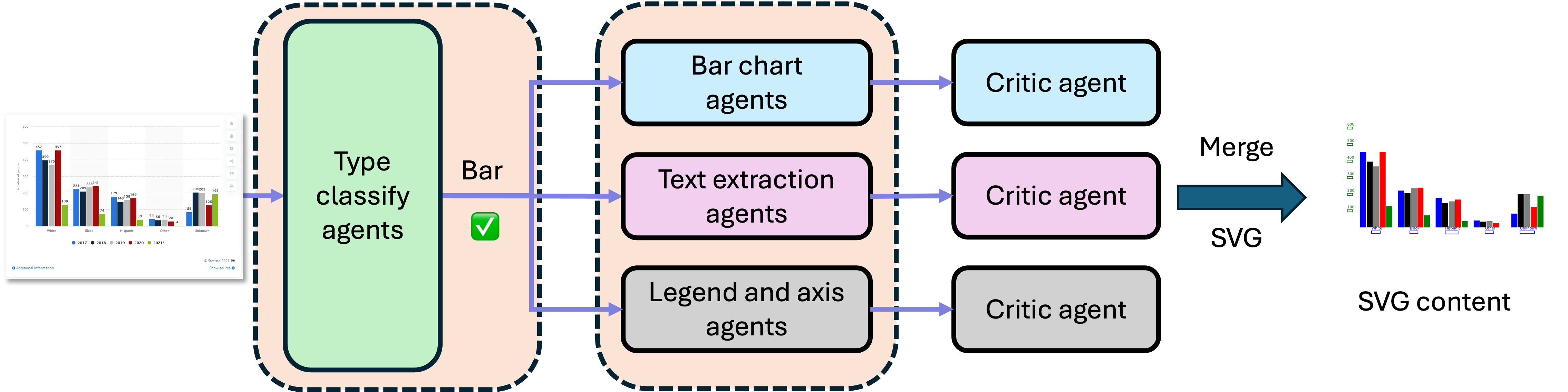}
\caption{Overview of the multi-agent collaboration pipeline for transforming chart images into Scalable Vector Graphics (SVG) representations. The process begins with chart type classification, followed by specialized agent-generators extracting semantic and geometric attributes (e.g., bar dimensions, text labels, legends). Agent-critics refine these outputs by identifying and correcting errors. The final merged SVG encodes all chart elements, enabling multimodal large language models (MLLMs) to perform robust zero-shot reasoning tasks like trend analysis and data interpretation.
  }
\label{fig:main}
\end{figure*}

\subsection{Multi-Agent Collaboration Pipeline for SVG Generation}

To generate high-quality SVG representations from chart images, we design a multi-agent collaboration pipeline that divides the task into specialized subtasks. 

\subsubsection{Chart Classification Agent}
Before activating specialized agents, the Chart Classification Agent first identifies the chart type, distinguishing between bar charts, line charts, and pie charts. This classification ensures that subsequent agents apply the most relevant extraction techniques tailored to the chart's structure to generate appropriate SVG elements.

\subsubsection{Agent-Generators for Specialized SVG Element Extraction}

Each agent-generator focuses on extracting specific attributes of the chart and converting them into corresponding SVG elements:

\begin{itemize}[nosep, leftmargin=0pt, topsep=0pt, partopsep=0pt]
    \item \textbf{Line Chart Agent:} Extracts line coordinates by identifying and mapping critical points along the chart's lines. Using detected key points, the agent generates a \texttt{<path>} element in the SVG, ensuring smooth transitions and preserving the geometric structure.
    \item \textbf{Bar Chart Agent:} Computes bar heights by processing bounding box dimensions for each detected bar. Each bar is represented as a \texttt{<rect>} element, with attributes such as \texttt{x}, \texttt{y}, \texttt{width}, and \texttt{height} correlating to the chart's data.
    \item \textbf{Pie Chart Agent:} Calculates proportions of color-coded segments, representing each segment as an arc in the SVG. This agent encodes angular or proportional values using \texttt{<path>} or similar elements to depict the pie chart.
    \item \textbf{Text Extraction Agent:} Associates textual content (e.g., axis labels, tick values, legend entries) with their spatial coordinates. The agent ensures that each text element is encoded as a \texttt{<text>} element, aligned with its corresponding visual component.
    \item \textbf{Legend and Axis Agent:} Focuses on semantic relationships, extracting and associating legend entries with their corresponding data representations and ensuring axis labels and ticks are mapped to their proper coordinates in the SVG.
\end{itemize}

The agent-generators operate in parallel, each producing a localized SVG component that aligns with the overall structure and semantics of the chart.
In practice, multiple agent-generators are deployed within each subtask, each using different tools to extract multiple versions of the results. 

\subsubsection{Agent-Critic for SVG Refinement}
In each subtask, agent-generators, such as the Line Chart Agent, are tasked with using different tools to extract multiple versions of the SVG elements. Once these agent-generators complete their subtasks, their outputs are passed to their respective agent-critics, each dedicated to evaluating a specific agent-generator. The agent-critics then engage in a deliberative evaluation, assessing the quality of the multiple results and determining how to refine the final SVG output by integrating insights from all versions. This process ensures that the symbolic representation is more accurate and consistent with the input chart. 

We prompt the agent-critic to identify inconsistencies or errors within the SVG elements, such as misaligned text, incorrect bounding box dimensions, or missing chart components. By systematically analyzing these discrepancies, the agent-critic will output the final refined SVG elements.
The agent-critic plays a particularly important role in handling complex charts with unusual layouts or styles, where errors in early stages of the pipeline could propagate and affect the final SVG output. By integrating insights from multiple generated versions, the agent-critic ensures that the final SVG representation is robust and reliable.

\subsection{Prompt Design of Agents for SVG Generation}

To effectively guide both the agent-generator and agent-critic in generating high-quality SVG elements, we design targeted prompts that direct their actions and decision-making processes. Below, we illustrate this approach using the Bar Chart Agent as an example of an agent-generator and its corresponding agent-critic. The full set of prompts can be found in the supplementary materials.  

\begin{itemize}[nosep, leftmargin=0pt, topsep=0pt, partopsep=0pt]
    \item \textbf{Prompt for Bar Chart Agent-Generator:} ``You are an expert in analyzing bar charts. Your task is to leverage available tools to extract key visual elements, specifically the bounding boxes of bars in the chart for SVG generation. Below are the tools at your disposal, all implemented in Python:  
def detection bbox1 (image, objects);  
def detection bbox2 (image, objects);  
...  
def detection bboxN (image, objects); 
Please indicate which tools you need to use to complete this task."
This prompt ensures that the agent-generator selects the most suitable tools to extract bounding box information, capturing essential bar elements for accurate SVG representation. 
\item \textbf{Prompt for Bar Chart Agent-Critic:}  
``You will be provided with a bar chart image and three generated SVG representations of the bounding boxes corresponding to bar elements in the chart. Your task is to:  
1. Analyze the three provided SVG codes, focusing on the bounding box elements representing the bars.  
2. Generate a final, comprehensive SVG code that consolidates information from all three SVG versions, ensuring it accurately captures all bar elements present in the original chart.  
3. Ensure that the final SVG representation correctly preserves bar positioning and the total number of bars."  
This structured prompt guides the agent-critic in evaluating inconsistencies, integrating insights from multiple generated versions, and producing a refined, high-fidelity SVG representation.  
\item \textbf{Prompt for MLLMs to Understand the Chart with SVG:} 
``You will be provided with a bar chart, its corresponding converted SVG representation, and a query. Your task is to answer the query while structuring your response into the following three fields:  
- instruction explanation: Explain the process of interpreting the chart based on its type and SVG structure.  
- explanation: Describe how you arrived at the answer using the provided SVG representation.  
- answer: Provide the final response based on the bar chart and the given query.''  
This structured output ensures that MLLMs can systematically process the symbolic SVG representation, apply reasoning to extract meaningful insights, and deliver an interpretable response.
\end{itemize}

\subsection{Details of Agent Tools for SVG Element Extraction}

The SVG-Chart pipeline processes a chart image $I$ through several stages, all aimed at generating an accurate SVG representation:

\begin{itemize}[nosep, leftmargin=0pt, topsep=0pt, partopsep=0pt]
    \item \textbf{Chart Type and Color Identification}
The chart image $I$ is input to a VLLM (e.g., GPT-4V) with a prompt to identify the chart type and the colors representing different data series.
This semantic understanding, obtained without task-specific fine-tuning, informs the subsequent SVG element generation process. 
\item \textbf{Color-Based Segmentation and Preprocessing}
The image is converted to HSV color space, where thresholds for hue, saturation, and value isolate elements of interest. A binary mask is created using 
\(\text{mask} = \text{inRange}(\text{HSV}, (H_{\min}, S_{\min}, V_{\min}), (H_{\max}, S_{\max}, V_{\max}))\). Refinement steps, such as morphological operations and Gaussian blurring, are applied to enhance the mask. For multi-series charts, this process is repeated for each identified color.
For bar charts, contours of segmented regions yield bounding boxes, encoded as \texttt{<rect>} elements in SVG. For line charts, key points along lines are detected using edge and peak detection algorithms, which are connected as \texttt{<path>} elements. For pie charts, proportions of segmented areas directly represent data ratios, encoded as colored SVG segments.
\item \textbf{Layout and Structure Detection}
Chart layout tools (e.g., ChartDete~\cite{yan2023context}) separate key regions, such as the plot area and legend. This segmentation ensures accurate extraction of data points, axes, and legend elements, which are processed according to their specific roles in the SVG representation.
\item \textbf{Text Recognition and Positioning}
OCR extracts textual content, such as axis labels, tick values, and legend entries. Each text element is associated with its bounding box coordinates $(x, y, w, h)$, ensuring precise alignment with chart elements. These are encoded as \texttt{<text>} elements in the SVG, maintaining semantic relationships between text and visual components.
\item \textbf{SVG Integration}
All extracted elements are compiled into a cohesive SVG document, preserving spatial relationships, semantic integrity, and quantitative data. This structured representation enables robust chart analysis by MLLMs across diverse chart types while maintaining both textual and geometric fidelity.
\end{itemize}

\section{Experiments}
\label{sec:experiment}

\subsection{Datasets}
We evaluate the effectiveness of our Socratic-Chart approach on ChartQA, including human-curated and augmented datasets. The charts cover various chart types such as pie, bar, and line charts. 

\subsection{Baselines and Metrics}
We compare our method against a broad set of baseline models, categorized as follows:

\begin{itemize}[nosep, leftmargin=0pt, topsep=0pt, partopsep=0pt]
    \item \textbf{Vision-Language Pre-trained (VLP) Models:} TaPas~\cite{holmgren2012tapas}, V-TaPas~\cite{masry2022chartqa}, T5~\cite{bujard198726}, VL-T5~\cite{cho2021unifying}, PaLI~\cite{chen2022pali}, Mini-GPT~\cite{zhu2023minigpt}, LLaVa~\cite{liu2024visual}, and GPT-4V~\cite{achiam2023gpt}. These models leverage large-scale pretraining but do not inherently use structured representations for charts.
    
    \item \textbf{Supervised Models:} ChartQA~\cite{masry2022chartqa}, ChartT5~\cite{zhou2023enhanced}, Pix2Struct~\cite{lee2023pix2struct}, MatCha~\cite{liu2022matcha}, Unichart~\cite{masry2023unichart}, and ChartLlama~\cite{han2023chartllama}. These models are specifically fine-tuned for chart-related tasks and often require extensive annotated datasets.
    
    \item \textbf{Table-Based Methods:} Deplot~\cite{liu2022deplot}, a table-augmented version of Unichart~\cite{masry2023unichart} and SIMPLOT~\cite{kim2024simplot}. These methods convert charts into tabular formats for reasoning in GPT-4V but may fail to capture important visual details.
\end{itemize}

Our method utilizes GPT-4V as the MLLM for the reported results.



In terms of evaluation metric, we use Relaxed Accuracy (RA), which measures the accuracy of question answering on the ChartQA dataset, considering both human-authored and augmented QA pairs. This is the standard evaluation metric adopted by previous chart understanding works, enabling direct and fair comparison across different approaches.







\begin{table}[ht]
\centering
\small 
\begin{tabular}{lccc}
\toprule
\textbf{Models} & \textbf{Human} & \textbf{Augmented} & \textbf{Overall} \\
\midrule
\midrule
\multicolumn{4}{l}{\textbf{VLP models}} \\
TaPas & 28.72 & 53.84 & 41.28 \\
V-TaPas & 29.60 & 61.44 & 45.52 \\
T5 & 25.12 & 56.96 & 41.04 \\
VL-T5 & 26.24 & 56.88 & 41.56 \\
PaLI & 30.40 & 64.90 & 47.65 \\
Mini-GPT & 8.40 & 15.60 & 12.00 \\
LLava & 37.68 & 72.96 & 55.32 \\
GPT-4 & 56.48 & 63.04 & 59.76 \\
\midrule
\multicolumn{4}{l}{\textbf{Supervised}} \\
ChartQA & 40.08 & 63.60 & 51.84 \\
ChartT5 & 31.80 & 74.40 & 53.10 \\
Pix2Struct & 30.50 & 81.60 & 56.05 \\
MatCha & 38.20 & 90.20 & 64.20 \\
Unichart & 43.92 & 88.56 & 66.24 \\
ChartLlama & 48.96 & \textbf{90.36} & 69.66 \\
\midrule
\textbf{Table} & & & \\
Deplot & 62.71 & 78.63 & 70.67 \\
Unichart\(^1\) & 67.04 & 69.92 & 68.48 \\
SIMPLOT & 78.07 & 88.42 & \textbf{83.24} \\
\midrule
\textbf{Socratic Chart (ours)} & \textbf{80.40} & 82.61 & 81.50 \\
\bottomrule
\end{tabular}
\caption{Chart question answering performance (RA) on the ChartQA dataset. Our method achieves strong performances over \textbf{Human} category, and reasonable competitive performances over \textbf{Augmented} category, resulting in 2nd best performanve in Overall. 
However, in ChartQA
augmented questions are programmatically generated with standardized, simple patterns that require less structural understanding, limiting the advantages of our SVG representation; in contrast, human-created questions feature natural complexity that benefits from the rich structural information our approach provides, better reflecting real-world applications. 
Therefore, these results show strong indications that our method captures the semantics of the charts.}
\label{tab:vlp_results}
\end{table}

\begin{table}[ht]
\centering
\small 
\begin{tabular}{lcccc}
\toprule
\textbf{Models} & \textbf{Human} & \textbf{Aug} & \textbf{Overall} & \textbf{Drop}\\
\midrule
\midrule
GPT-4v &  33.5 &  48.0 &  40.8 &  37.6\\
GPT-4o &  47.6 &  52.4 &  50.0 &  35.7\\
Gemini2 & 31.6 &  42.8 &  37.2 &  50.0\\
SIMPLOT &  38.0  &  55.3   &  46.7 &  36.5\\
\textbf{Socratic Chart w SA} &  50.4  &  60.1 & 55.3 &  24.2\\
\textbf{Socratic Chart w MA } &  \textbf{52.6} &  \textbf{62.5}  &  \textbf{57.6} &  \textbf{23.9}\\
\bottomrule
\end{tabular}
\caption{Chart question answering performance (RA) on the \textbf{ChartQA-RL} (ChartQA with removed labels). SA stands for single agent, MA stands for multiple agent. The Drop column shows percentage performance decrease compared to standard ChartQA. Our approach demonstrates superior robustness to label removal with the lowest performance drop (23.9\%).
}
\label{rl_results}
\end{table}

\begin{table}[ht]
\centering
\small 
\begin{tabular}{lcccc}
\toprule
\textbf{Models} & \textbf{Human} & \textbf{Aug} & \textbf{Overall} & \textbf{Drop}\\
\midrule
\midrule
GPT-4v &  50.2 &  66.8 &  58.5 &  19.9\\
GPT-4o &  67.0 &  76.0 &  71.5 &  14.2\\
Gemini2 &  47.4 &  57.8 &  52.6 &  34.6\\
SIMPLOT & 51.7  &  73.8 &  62.8 &  20.4\\
\textbf{Socratic Chart w SA} & 58.8 & 75.0 & 66.9 &  \textbf{12.6}\\
\textbf{Socratic Chart w MA } & \textbf{60.1}  & \textbf{76.1} & \textbf{68.1} & 13.4\\
\bottomrule
\end{tabular}
\caption{Performance (RA) on \textbf{ChartQA-HV} (ChartQA with horizontal and vertical perturbation). SA stands for single agent, MA stands for multiple agent. The Drop column indicates performance decrease compared to standard ChartQA. Both our single and multi-agent approaches demonstrate exceptional resilience to visual perturbations, with the lowest performance drops (12.6\% and 13.4\%) among all models. }
\label{results_perturbation}
\end{table}

\begin{table}[ht]
\centering
\small 
\begin{tabular}{lcccc}
\toprule
\textbf{Models} & \textbf{Overall}\\
\midrule
\midrule
GPT-4v &  26.9\\
SIMPLOT & 14.1\\
\textbf{Socratic Chart w MA } & \textbf{30.1}\\
\bottomrule
\end{tabular}
\caption{Performance on \textbf{Charixv}. Our Socratic Chart with multiple agents (MA) achieves the highest overall accuracy (30.1\%), significantly outperforming both GPT-4v (26.9\%) and SIMPLOT (11.1\%).}
\label{results_charixv}
\end{table}

\begin{figure*}[t]
\centering
\includegraphics[width=1.0\textwidth]{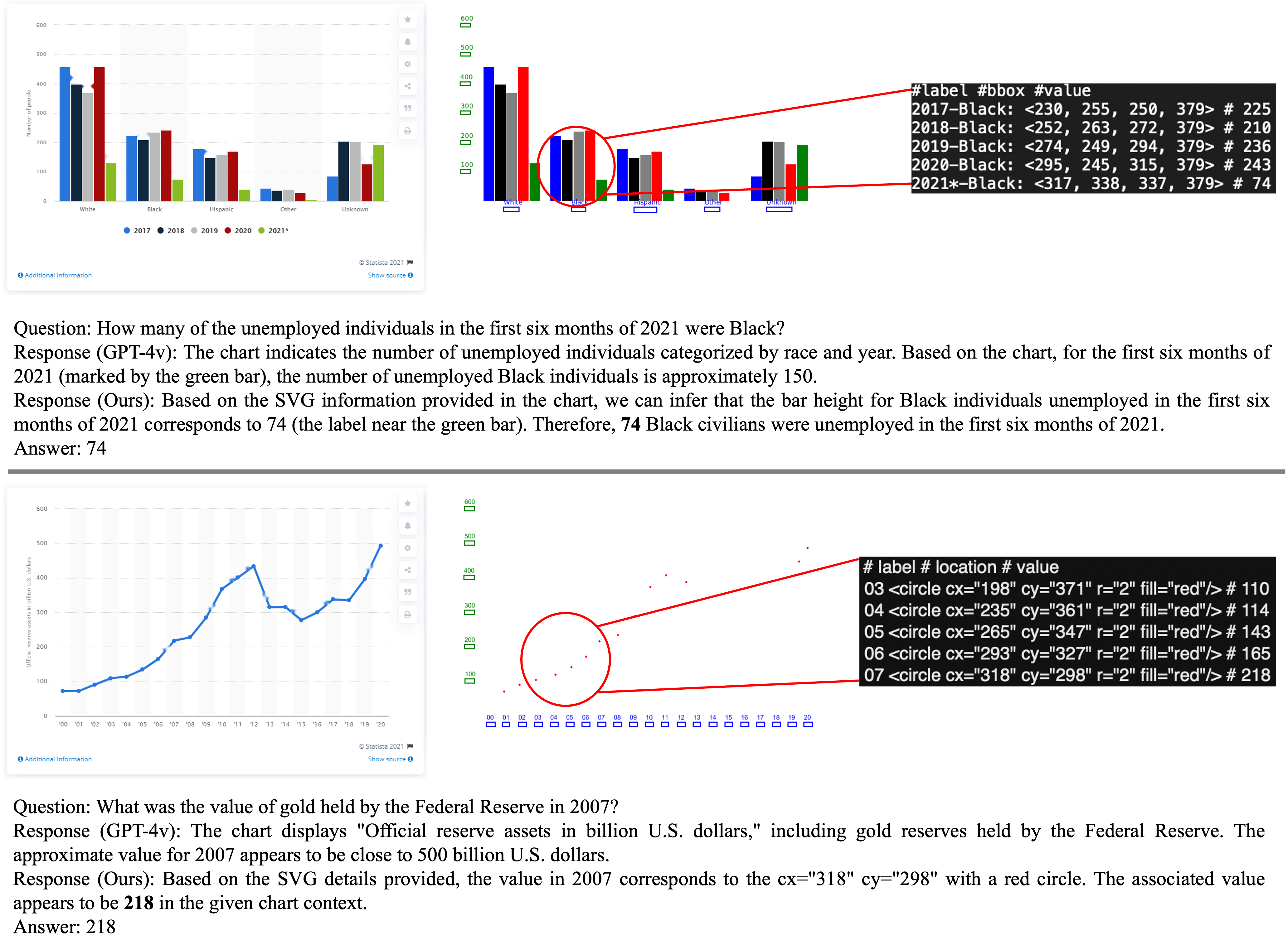}
\vspace{-1.5em}
\caption{We present a visualization of the SVG code generated using our multi-agent pipeline, demonstrating its ability to assist MLLMs in answering questions that GPT-4v fails to address correctly. The SVG representation encodes the geometric and semantic attributes of the chart, such as bar heights, line coordinates, and text labels, in a structured format. This high-fidelity representation enables MLLMs to perform precise reasoning tasks, such as extracting specific data points or identifying trends, even in complex charts where traditional models like GPT-4v may struggle due to their reliance on visual estimation. By leveraging the symbolic abstraction provided by the SVG format, our framework ensures accurate and interpretable chart understanding.
}
\label{fig:svg_example1}
\end{figure*}




\subsection{Evaluation on ChartQA}

To assess the effectiveness of our framework, we conducted extensive evaluations on ChartQA and related OOD benchmarks designed for chart reasoning tasks.
Our results demonstrate that the proposed framework, \textit{Socratic Chart}, outperforms state-of-the-art vision-language models, including GPT-4V and Gemini-2.0 Pro, across a wide range of metrics.

\subsubsection{Quantitative Performance}

In addition to quantitative improvements, qualitative evaluations highlight the robustness of our framework in understanding and reasoning over complex charts. For example, our method demonstrates superior performance in handling challenging scenarios such as label removal and perturbation, as evidenced by the results in Tables~\ref{rl_results} and~\ref{results_perturbation}.

\begin{itemize}[nosep, leftmargin=0pt, topsep=0pt, partopsep=0pt]
\item \textbf{Label Removal:} On the ChartQA dataset with removed labels (we name it \textbf{ChartQA-RL}), our method (Socratic-Chart with multiple agents, MA) achieves an overall accuracy of 57.6\%, significantly outperforming GPT-4V (40.8\%) and Gemini2 (37.2\%). This indicates that our framework can effectively infer missing information by leveraging the structured SVG representation and multi-agent collaboration, even when critical textual cues are absent.

\item \textbf{Perturbation:} When tested on the ChartQA dataset with horizontal and vertical perturbations (we name it \textbf{ChartQA-HV}), Socratic-Chart with MA achieves an overall accuracy of 68.1\%, surpassing GPT-4V (58.5\%) and SIMPLOT (62.8\%). This demonstrates the robustness of our approach in handling distorted or misaligned chart elements, which are common in real-world scenarios.

\item \textbf{Multi-Agent Collaboration:} The consistent improvement of Socratic-Chart with multiple agents (MA) over the single-agent (SA) variant across both datasets highlights the importance of specialized agent-generators. For instance, in the label removal task, MA improves over SA by 2.2\% in overall accuracy, while in the perturbation task, the improvement is 1.2\%. This suggests that the division of labor among agents enhances the model's ability to extract and reason over chart attributes accurately. Additionally, the multi-agent framework allows for the use of a critic agent, which facilitates the discussion and synthesis of results from different agents. This collaborative process further refines the final output, ensuring higher accuracy and robustness in handling complex tasks.

\end{itemize}

These qualitative insights, combined with the quantitative results, demonstrate that our framework excels in handling complex and noisy chart data, making it a robust solution for real-world chart understanding tasks.


    

\subsubsection{Qualitative Analysis}
In Figure~\ref{fig:svg_example1}, We present a detailed visualization of the SVG code generated using our multi-agent pipeline, showcasing how it effectively encodes the semantic, textual, and geometric elements of charts. This structured representation not only ensures accurate chart reconstruction but also enhances the ability of MLLMs to interpret and reason about visual data. By leveraging this high-fidelity SVG output, MLLMs can answer complex questions that GPT-4v fails to address correctly, particularly those requiring a deeper integration of textual and visual modalities. 

\subsubsection{Evaluation on Charixv}
To further evaluate our framework's generalization capabilities, we conducted experiments on Charixv, a challenging benchmark featuring complex chart types and diverse reasoning tasks. Unlike ChartQA, Charixv includes unconventional layouts and visualization styles that better represent real-world variability in chart design. Table~\ref{results_charixv} shows our method's performance on this challenging dataset.

\subsection{Ablation Studies}

\paragraph{Contribution of Agent Components} To evaluate the contribution of individual components, we conducted ablation studies by selectively disabling certain aspects of the pipeline in Table~\ref{tab:ablation}:

\begin{table}[ht]
\centering
\small
\begin{tabular}{lccc}
\toprule
\textbf{Ablation} & \textbf{Human} & \textbf{Aug} & \textbf{Overall}\\
\midrule
\midrule
Socratic Chart w MA &  60.1 &  76.1 &  68.1\\
\midrule
Without Agent-Critic & 58.8 & 75.0  & 66.9 \\
Without Specialized Agents & 56.4 & 73.2 & 64.8 \\
\bottomrule
\end{tabular}
\caption{Ablation study of SVG-Chart on the ChartQA dataset with horizontal and vertical perturbation. The table shows the performance (Relaxed Accuracy) when removing key components of the pipeline: Agent-Critic, Specialized Agents, and SVG Integration. MA stands for multiple agents.}
\label{tab:ablation}
\end{table}

\begin{itemize}[nosep, leftmargin=0pt, topsep=0pt, partopsep=0pt]
    \item \textbf{Without Agent-Critic:} Removing the agent-critic resulted in a noticeable drop in accuracy, particularly in multi-step reasoning tasks. This highlights the importance of the validation and refinement stage for ensuring high-fidelity representations.
    \item \textbf{Without Specialized Agent-Generators:} Replacing specialized agent-generators with a single general-purpose extractor led to reduced performance.
\end{itemize}

\paragraph{Contribution of SVG Components}
In the second ablation experiment, we investigated the impact of specific SVG components in Table~\ref{tab:ablation2}. This experiment was designed to quantify the contribution of individual structural elements within the SVG representation, helping us understand which components are most critical. 

\begin{table}[ht]
\centering
\small
\begin{tabular}{lccc}
\toprule
\textbf{Ablation} & \textbf{Human} & \textbf{Aug} & \textbf{Overall}\\
\midrule
\midrule
Socratic Chart w MA &  60.1 &  76.1 &  68.1\\
\midrule
w/o x-axis  & 58.1 & 75.5  & 66.8 \\
w/o y-axis & 51.5 & 70.3 & 60.9 \\
w/o x and y-axis & 50.6 & 67.7 & 59.2 \\
\bottomrule
\end{tabular}
\caption{Ablation study of SVG-Chart on the ChartQA dataset with horizontal and vertical perturbation. The table shows the performance (Relaxed Accuracy) when removing key SVG components x-axis or y-axis.}
\label{tab:ablation2}
\end{table}

\begin{itemize}[nosep, leftmargin=0pt, topsep=0pt, partopsep=0pt]
    \item \textbf{Without X-axis Elements:} Removing x-axis elements from the SVG parsing resulted in a relatively minor performance decrease (-1.3\% overall), suggesting that while horizontal anchoring is important, the model can often infer this information from other chart elements.
    \item \textbf{Without Y-axis Elements:} The removal of y-axis elements had a much more significant impact (-7.2\% overall), particularly on human-created set (-8.6\%). This highlights the critical importance of vertical scale information for accurate quantitative reasoning.
    \item \textbf{Without Both X and Y Axes:} When both axes were removed, performance declined dramatically (-8.9\% overall), with comparable degradation across both human-created (-9.5\%) and augmented set (-8.4\%). 
\end{itemize}









\subsection{Limitations}
Despite the promising results of Socratic Chart, several limitations remain in the current framework. The multi-agent pipeline introduces computational overhead that may impact real-time applications, 
the SVG generation process requires multiple specialized agents working in sequence, which increases the overall inference time and our method still does not have perfect performance even though it is better than SOTA ones on human set. 
\section{Conclusion}
\label{sec:conclusioin}

Multimodal Large Language Models MLLMs show impressive capabilities but struggle with genuine visual reasoning in chart understanding. Enhanced evaluations on the ChartQA benchmark, including removing text labels and introducing perturbations, revealed heavy reliance on OCR-based shortcuts, with performance drops of up to 30\% in models like GPT-4o and Gemini-2.0 Pro.

To address this, we proposed Socratic Chart, a framework that converts charts into Scalable Vector Graphics (SVG) representations, enabling MLLMs to integrate textual and visual modalities. Using a multi-agent pipeline with specialized agent-generators and an agent-critic for refinement, Socratic Chart ensures high-fidelity symbolic representations. It outperforms state-of-the-art models, demonstrating superior robustness and reasoning capabilities under challenging conditions, advancing the path toward true multimodal understanding.

\small
\bibliographystyle{ieeenat_fullname}
\bibliography{main}

\begin{thebibliography}{42}
\providecommand{\natexlab}[1]{#1}
\providecommand{\url}[1]{\texttt{#1}}
\expandafter\ifx\csname urlstyle\endcsname\relax
  \providecommand{\doi}[1]{doi: #1}\else
  \providecommand{\doi}{doi: \begingroup \urlstyle{rm}\Url}\fi

\bibitem[Achiam et~al.(2023)Achiam, Adler, Agarwal, Ahmad, Akkaya, Aleman, Almeida, Altenschmidt, Altman, Anadkat, et~al.]{achiam2023gpt}
Josh Achiam, Steven Adler, Sandhini Agarwal, Lama Ahmad, Ilge Akkaya, Florencia~Leoni Aleman, Diogo Almeida, Janko Altenschmidt, Sam Altman, Shyamal Anadkat, et~al.
\newblock Gpt-4 technical report.
\newblock \emph{arXiv preprint arXiv:2303.08774}, 2023.

\bibitem[Alayrac et~al.(2022)Alayrac, Donahue, Luc, Miech, Barr, Hasson, Lenc, Mensch, Millican, Reynolds, et~al.]{alayrac2022flamingo}
Jean-Baptiste Alayrac, Jeff Donahue, Pauline Luc, Antoine Miech, Iain Barr, Yana Hasson, Karel Lenc, Arthur Mensch, Katherine Millican, Malcolm Reynolds, et~al.
\newblock Flamingo: a visual language model for few-shot learning.
\newblock \emph{Advances in neural information processing systems}, 35:\penalty0 23716--23736, 2022.

\bibitem[Antol et~al.(2015)Antol, Agrawal, Lu, Mitchell, Batra, Zitnick, and Parikh]{antol2015vqa}
Stanislaw Antol, Aishwarya Agrawal, Jiasen Lu, Margaret Mitchell, Dhruv Batra, C~Lawrence Zitnick, and Devi Parikh.
\newblock Vqa: Visual question answering.
\newblock In \emph{Proceedings of the IEEE international conference on computer vision}, pages 2425--2433, 2015.

\bibitem[Bai et~al.(2023)Bai, Bai, Yang, Wang, Tan, Wang, Lin, Zhou, and Zhou]{bai2023qwen}
Jinze Bai, Shuai Bai, Shusheng Yang, Shijie Wang, Sinan Tan, Peng Wang, Junyang Lin, Chang Zhou, and Jingren Zhou.
\newblock Qwen-vl: A versatile vision-language model for understanding, localization, text reading, and beyond.
\newblock \emph{arXiv preprint arXiv:2308.12966}, 1\penalty0 (2):\penalty0 3, 2023.

\bibitem[Bujard et~al.(1987)Bujard, Gentz, Lanzer, Stueber, Mueller, Ibrahimi, Haeuptle, and Dobberstein]{bujard198726}
Hermann Bujard, Reiner Gentz, Michael Lanzer, Dietrich Stueber, Michael Mueller, Ibrahim Ibrahimi, Marie-Therese Haeuptle, and Bernhard Dobberstein.
\newblock [26] a t5 promoter-based transcription-translation system for the analysis of proteins in vitro and in vivo.
\newblock In \emph{Methods in enzymology}, pages 416--433. Elsevier, 1987.

\bibitem[Chen et~al.(2024{\natexlab{a}})Chen, Zhang, Zhou, Rossi, Gu, and Chen]{chen2024mmr}
Jian Chen, Ruiyi Zhang, Yufan Zhou, Ryan Rossi, Jiuxiang Gu, and Changyou Chen.
\newblock Mmr: Evaluating reading ability of large multimodal models.
\newblock \emph{arXiv preprint arXiv:2408.14594}, 2024{\natexlab{a}}.

\bibitem[Chen et~al.(2024{\natexlab{b}})Chen, Li, Dong, Zhang, Zang, Chen, Duan, Wang, Qiao, Lin, et~al.]{chen2024we}
Lin Chen, Jinsong Li, Xiaoyi Dong, Pan Zhang, Yuhang Zang, Zehui Chen, Haodong Duan, Jiaqi Wang, Yu Qiao, Dahua Lin, et~al.
\newblock Are we on the right way for evaluating large vision-language models?
\newblock \emph{arXiv preprint arXiv:2403.20330}, 2024{\natexlab{b}}.

\bibitem[Chen et~al.(2022)Chen, Wang, Changpinyo, Piergiovanni, Padlewski, Salz, Goodman, Grycner, Mustafa, Beyer, et~al.]{chen2022pali}
Xi Chen, Xiao Wang, Soravit Changpinyo, AJ Piergiovanni, Piotr Padlewski, Daniel Salz, Sebastian Goodman, Adam Grycner, Basil Mustafa, Lucas Beyer, et~al.
\newblock Pali: A jointly-scaled multilingual language-image model.
\newblock \emph{arXiv preprint arXiv:2209.06794}, 2022.

\bibitem[Cho et~al.(2021)Cho, Lei, Tan, and Bansal]{cho2021unifying}
Jaemin Cho, Jie Lei, Hao Tan, and Mohit Bansal.
\newblock Unifying vision-and-language tasks via text generation.
\newblock In \emph{International Conference on Machine Learning}, pages 1931--1942. PMLR, 2021.

\bibitem[Ferraiolo et~al.(2000)Ferraiolo, Jun, and Jackson]{ferraiolo2000scalable}
Jon Ferraiolo, Fujisawa Jun, and Dean Jackson.
\newblock \emph{Scalable vector graphics (SVG) 1.0 specification}.
\newblock iuniverse Bloomington, 2000.

\bibitem[Fu et~al.(2024)Fu, Hu, Li, Feng, Wang, Lin, Roth, Smith, Ma, and Krishna]{fu2024blink}
Xingyu Fu, Yushi Hu, Bangzheng Li, Yu Feng, Haoyu Wang, Xudong Lin, Dan Roth, Noah~A Smith, Wei-Chiu Ma, and Ranjay Krishna.
\newblock Blink: Multimodal large language models can see but not perceive.
\newblock In \emph{European Conference on Computer Vision}, pages 148--166. Springer, 2024.

\bibitem[Guan et~al.(2024)Guan, Liu, Wu, Xian, Li, Liu, Wang, Chen, Huang, Yacoob, et~al.]{guan2024hallusionbench}
Tianrui Guan, Fuxiao Liu, Xiyang Wu, Ruiqi Xian, Zongxia Li, Xiaoyu Liu, Xijun Wang, Lichang Chen, Furong Huang, Yaser Yacoob, et~al.
\newblock Hallusionbench: an advanced diagnostic suite for entangled language hallucination and visual illusion in large vision-language models.
\newblock In \emph{Proceedings of the IEEE/CVF Conference on Computer Vision and Pattern Recognition}, pages 14375--14385, 2024.

\bibitem[Gupta and Kembhavi(2023)]{gupta2023visual}
Tanmay Gupta and Aniruddha Kembhavi.
\newblock Visual programming: Compositional visual reasoning without training.
\newblock In \emph{Proceedings of the IEEE/CVF Conference on Computer Vision and Pattern Recognition}, pages 14953--14962, 2023.

\bibitem[Han et~al.(2023)Han, Zhang, Chen, Yang, Wang, Yu, Fu, and Zhang]{han2023chartllama}
Yucheng Han, Chi Zhang, Xin Chen, Xu Yang, Zhibin Wang, Gang Yu, Bin Fu, and Hanwang Zhang.
\newblock Chartllama: A multimodal llm for chart understanding and generation.
\newblock \emph{arXiv preprint arXiv:2311.16483}, 2023.

\bibitem[Hegde et~al.(2023)Hegde, Paul, Madan, and Aggarwal]{hegde2023analyzing}
Nidhi Hegde, Sujoy Paul, Gagan Madan, and Gaurav Aggarwal.
\newblock Analyzing the efficacy of an llm-only approach for image-based document question answering.
\newblock \emph{arXiv preprint arXiv:2309.14389}, 2023.

\bibitem[Holmgren et~al.(2012)Holmgren, Davidsson, Persson, and Ramstedt]{holmgren2012tapas}
Johan Holmgren, Paul Davidsson, Jan~A Persson, and Linda Ramstedt.
\newblock Tapas: A multi-agent-based model for simulation of transport chains.
\newblock \emph{Simulation Modelling Practice and Theory}, 23:\penalty0 1--18, 2012.

\bibitem[Hu et~al.(2022)Hu, Hua, Yang, Shi, Smith, and Luo]{hu2022promptcap}
Yushi Hu, Hang Hua, Zhengyuan Yang, Weijia Shi, Noah~A Smith, and Jiebo Luo.
\newblock Promptcap: Prompt-guided task-aware image captioning.
\newblock \emph{arXiv preprint arXiv:2211.09699}, 2022.

\bibitem[Hu et~al.(2024)Hu, Stretcu, Lu, Viswanathan, Hata, Luo, Krishna, and Fuxman]{hu2024visual}
Yushi Hu, Otilia Stretcu, Chun-Ta Lu, Krishnamurthy Viswanathan, Kenji Hata, Enming Luo, Ranjay Krishna, and Ariel Fuxman.
\newblock Visual program distillation: Distilling tools and programmatic reasoning into vision-language models.
\newblock In \emph{Proceedings of the IEEE/CVF Conference on Computer Vision and Pattern Recognition}, pages 9590--9601, 2024.

\bibitem[Jain et~al.(2024)Jain, Yang, and Shi]{jain2024vcoder}
Jitesh Jain, Jianwei Yang, and Humphrey Shi.
\newblock Vcoder: Versatile vision encoders for multimodal large language models.
\newblock In \emph{Proceedings of the IEEE/CVF Conference on Computer Vision and Pattern Recognition}, pages 27992--28002, 2024.

\bibitem[Jiang et~al.(2024)Jiang, Xie, Hao, Wang, Mallick, Su, Taylor, and Roth]{jiang2024peek}
Bowen Jiang, Yangxinyu Xie, Zhuoqun Hao, Xiaomeng Wang, Tanwi Mallick, Weijie~J Su, Camillo~J Taylor, and Dan Roth.
\newblock A peek into token bias: Large language models are not yet genuine reasoners.
\newblock \emph{arXiv preprint arXiv:2406.11050}, 2024.

\bibitem[Kim et~al.(2024)Kim, Park, In, Han, and Park]{kim2024simplot}
Wonjoong Kim, Sangwu Park, Yeonjun In, Seokwon Han, and Chanyoung Park.
\newblock Simplot: Enhancing chart question answering by distilling essentials.
\newblock \emph{arXiv preprint arXiv:2405.00021}, 2024.

\bibitem[Lee et~al.(2023)Lee, Joshi, Turc, Hu, Liu, Eisenschlos, Khandelwal, Shaw, Chang, and Toutanova]{lee2023pix2struct}
Kenton Lee, Mandar Joshi, Iulia~Raluca Turc, Hexiang Hu, Fangyu Liu, Julian~Martin Eisenschlos, Urvashi Khandelwal, Peter Shaw, Ming-Wei Chang, and Kristina Toutanova.
\newblock Pix2struct: Screenshot parsing as pretraining for visual language understanding.
\newblock In \emph{International Conference on Machine Learning}, pages 18893--18912. PMLR, 2023.

\bibitem[Li et~al.(2024)Li, Ge, Ge, Wang, Wang, Zhang, and Shan]{li2024seed}
Bohao Li, Yuying Ge, Yixiao Ge, Guangzhi Wang, Rui Wang, Ruimao Zhang, and Ying Shan.
\newblock Seed-bench: Benchmarking multimodal large language models.
\newblock In \emph{Proceedings of the IEEE/CVF Conference on Computer Vision and Pattern Recognition}, pages 13299--13308, 2024.

\bibitem[Liu et~al.(2022{\natexlab{a}})Liu, Eisenschlos, Piccinno, Krichene, Pang, Lee, Joshi, Chen, Collier, and Altun]{liu2022deplot}
Fangyu Liu, Julian~Martin Eisenschlos, Francesco Piccinno, Syrine Krichene, Chenxi Pang, Kenton Lee, Mandar Joshi, Wenhu Chen, Nigel Collier, and Yasemin Altun.
\newblock Deplot: One-shot visual language reasoning by plot-to-table translation.
\newblock \emph{arXiv preprint arXiv:2212.10505}, 2022{\natexlab{a}}.

\bibitem[Liu et~al.(2022{\natexlab{b}})Liu, Piccinno, Krichene, Pang, Lee, Joshi, Altun, Collier, and Eisenschlos]{liu2022matcha}
Fangyu Liu, Francesco Piccinno, Syrine Krichene, Chenxi Pang, Kenton Lee, Mandar Joshi, Yasemin Altun, Nigel Collier, and Julian~Martin Eisenschlos.
\newblock Matcha: Enhancing visual language pretraining with math reasoning and chart derendering.
\newblock \emph{arXiv preprint arXiv:2212.09662}, 2022{\natexlab{b}}.

\bibitem[Liu et~al.(2024)Liu, Li, Wu, and Lee]{liu2024visual}
Haotian Liu, Chunyuan Li, Qingyang Wu, and Yong~Jae Lee.
\newblock Visual instruction tuning.
\newblock \emph{Advances in neural information processing systems}, 36, 2024.

\bibitem[Masry et~al.(2022)Masry, Long, Tan, Joty, and Hoque]{masry2022chartqa}
Ahmed Masry, Do~Xuan Long, Jia~Qing Tan, Shafiq Joty, and Enamul Hoque.
\newblock Chartqa: A benchmark for question answering about charts with visual and logical reasoning.
\newblock \emph{arXiv preprint arXiv:2203.10244}, 2022.

\bibitem[Masry et~al.(2023)Masry, Kavehzadeh, Do, Hoque, and Joty]{masry2023unichart}
Ahmed Masry, Parsa Kavehzadeh, Xuan~Long Do, Enamul Hoque, and Shafiq Joty.
\newblock Unichart: A universal vision-language pretrained model for chart comprehension and reasoning.
\newblock \emph{arXiv preprint arXiv:2305.14761}, 2023.

\bibitem[Nasiriany et~al.(2024)Nasiriany, Xia, Yu, Xiao, Liang, Dasgupta, Xie, Driess, Wahid, Xu, et~al.]{nasiriany2024pivot}
Soroush Nasiriany, Fei Xia, Wenhao Yu, Ted Xiao, Jacky Liang, Ishita Dasgupta, Annie Xie, Danny Driess, Ayzaan Wahid, Zhuo Xu, et~al.
\newblock Pivot: Iterative visual prompting elicits actionable knowledge for vlms.
\newblock \emph{arXiv preprint arXiv:2402.07872}, 2024.

\bibitem[Peng and Zhang(2004)]{peng2004roles}
Zhong-Ren Peng and Chuanrong Zhang.
\newblock The roles of geography markup language (gml), scalable vector graphics (svg), and web feature service (wfs) specifications in the development of internet geographic information systems (gis).
\newblock \emph{Journal of Geographical Systems}, 6:\penalty0 95--116, 2004.

\bibitem[Rahmanzadehgervi et~al.(2024)Rahmanzadehgervi, Bolton, Taesiri, and Nguyen]{rahmanzadehgervi2024vision}
Pooyan Rahmanzadehgervi, Logan Bolton, Mohammad~Reza Taesiri, and Anh~Totti Nguyen.
\newblock Vision language models are blind.
\newblock In \emph{Proceedings of the Asian Conference on Computer Vision}, pages 18--34, 2024.

\bibitem[Sur{\'\i}s et~al.(2023)Sur{\'\i}s, Menon, and Vondrick]{suris2023vipergpt}
D{\'\i}dac Sur{\'\i}s, Sachit Menon, and Carl Vondrick.
\newblock Vipergpt: Visual inference via python execution for reasoning.
\newblock In \emph{Proceedings of the IEEE/CVF International Conference on Computer Vision}, pages 11888--11898, 2023.

\bibitem[Team et~al.(2023)Team, Anil, Borgeaud, Alayrac, Yu, Soricut, Schalkwyk, Dai, Hauth, Millican, et~al.]{team2023gemini}
Gemini Team, Rohan Anil, Sebastian Borgeaud, Jean-Baptiste Alayrac, Jiahui Yu, Radu Soricut, Johan Schalkwyk, Andrew~M Dai, Anja Hauth, Katie Millican, et~al.
\newblock Gemini: a family of highly capable multimodal models.
\newblock \emph{arXiv preprint arXiv:2312.11805}, 2023.

\bibitem[Wang et~al.(2024)Wang, Xia, He, Chen, Liu, Zhu, Liang, Wu, Liu, Malladi, et~al.]{wang2024charxiv}
Zirui Wang, Mengzhou Xia, Luxi He, Howard Chen, Yitao Liu, Richard Zhu, Kaiqu Liang, Xindi Wu, Haotian Liu, Sadhika Malladi, et~al.
\newblock Charxiv: Charting gaps in realistic chart understanding in multimodal llms.
\newblock \emph{Advances in Neural Information Processing Systems}, 37:\penalty0 113569--113697, 2024.

\bibitem[Yan et~al.(2023{\natexlab{a}})Yan, Yang, Zhu, Lin, Li, Wang, Yang, Zhong, McAuley, Gao, et~al.]{yan2023gpt}
An Yan, Zhengyuan Yang, Wanrong Zhu, Kevin Lin, Linjie Li, Jianfeng Wang, Jianwei Yang, Yiwu Zhong, Julian McAuley, Jianfeng Gao, et~al.
\newblock Gpt-4v in wonderland: Large multimodal models for zero-shot smartphone gui navigation.
\newblock \emph{arXiv preprint arXiv:2311.07562}, 2023{\natexlab{a}}.

\bibitem[Yan et~al.(2023{\natexlab{b}})Yan, Ahmed, and Doermann]{yan2023context}
Pengyu Yan, Saleem Ahmed, and David Doermann.
\newblock Context-aware chart element detection.
\newblock In \emph{International conference on document analysis and recognition}, pages 218--233. Springer, 2023{\natexlab{b}}.

\bibitem[Yang et~al.(2023)Yang, Li, Wang, Lin, Azarnasab, Ahmed, Liu, Liu, Zeng, and Wang]{yang2023mm}
Zhengyuan Yang, Linjie Li, Jianfeng Wang, Kevin Lin, Ehsan Azarnasab, Faisal Ahmed, Zicheng Liu, Ce Liu, Michael Zeng, and Lijuan Wang.
\newblock Mm-react: Prompting chatgpt for multimodal reasoning and action.
\newblock \emph{arXiv preprint arXiv:2303.11381}, 2023.

\bibitem[Yao et~al.(2023)Yao, Zhao, Yu, Du, Shafran, Narasimhan, and Cao]{yao2023react}
Shunyu Yao, Jeffrey Zhao, Dian Yu, Nan Du, Izhak Shafran, Karthik Narasimhan, and Yuan Cao.
\newblock React: synergizing reasoning and acting in language models (2022).
\newblock \emph{arXiv preprint arXiv:2210.03629}, 2023.

\bibitem[Yasunaga et~al.(2022)Yasunaga, Aghajanyan, Shi, James, Leskovec, Liang, Lewis, Zettlemoyer, and Yih]{yasunaga2022retrieval}
Michihiro Yasunaga, Armen Aghajanyan, Weijia Shi, Rich James, Jure Leskovec, Percy Liang, Mike Lewis, Luke Zettlemoyer, and Wen-tau Yih.
\newblock Retrieval-augmented multimodal language modeling.
\newblock \emph{arXiv preprint arXiv:2211.12561}, 2022.

\bibitem[Zhang et~al.(2023)Zhang, Liu, Zhang, Cheng, and Wang]{zhang2023beyond}
Tong Zhang, Haoyang Liu, Peiyan Zhang, Yuxuan Cheng, and Haohan Wang.
\newblock Beyond pixels: Exploring human-readable svg generation for simple images with vision language models.
\newblock \emph{arXiv preprint arXiv:2311.15543}, 2023.

\bibitem[Zhou et~al.(2023)Zhou, Fung, Chen, Thomas, Ji, and Chang]{zhou2023enhanced}
Mingyang Zhou, Yi~R Fung, Long Chen, Christopher Thomas, Heng Ji, and Shih-Fu Chang.
\newblock Enhanced chart understanding in vision and language task via cross-modal pre-training on plot table pairs.
\newblock \emph{arXiv preprint arXiv:2305.18641}, 2023.

\bibitem[Zhu et~al.(2023)Zhu, Chen, Shen, Li, and Elhoseiny]{zhu2023minigpt}
Deyao Zhu, Jun Chen, Xiaoqian Shen, Xiang Li, and Mohamed Elhoseiny.
\newblock Minigpt-4: Enhancing vision-language understanding with advanced large language models.
\newblock \emph{arXiv preprint arXiv:2304.10592}, 2023.

\end{thebibliography}


\end{document}